\documentclass[runningheads]{llncs}
\usepackage[T1]{fontenc}
\usepackage{graphicx}
\usepackage{amsmath}
\usepackage{url}
\usepackage{booktabs}
\makeatletter
\@ifundefined{credits}{%
  \newenvironment{credits}{}{}%
  \providecommand{\discintname}{Disclosure of Interests.}%
}{}
\makeatother
\providecommand{\doi}[1]{\url{https://doi.org/#1}}
\usepackage{array}
\newcolumntype{P}[1]{>{\raggedright\arraybackslash}p{#1}}

\begin{document}
%

\title{BS: Take the Hint - Interactive Multitracer PET/CT Lesion
Segmentation with a Scribble-Conditioned ResEnc U-Net}

%
\titlerunning{Interactive Lesion Segmentation in Multitracer PET/CT}
%
\author{Marven Sherif\inst{1} \and
Amgad Elmasry\inst{1} \and
Youssef Ghazal\inst{1} \and
Ayman Elghotni\inst{1}}
\authorrunning{M. Sherif et al.}
%
\institute{Brightskies, Egypt\\
\email{\{marven.sherif,amgad.elmasry,youssef.ghazal,ayman.elghotni\}@brightskiesinc.com}}
\maketitle              
\begin{abstract}
Automated lesion segmentation in whole-body PET/CT is complicated by the
variety of physiological tracer uptake patterns and by the differing
appearance of lesions across tracers. The autoPET/CT V challenge addresses
this by making segmentation \emph{interactive}: user scribbles marking
foreground and background are supplied alongside the image, and the algorithm
is expected to exploit them. We present our submission, a
scribble-conditioned residual encoder U-Net operating on four input channels
-- CT, PET, and a sparse scribble map for each of foreground and background. The network is initialised from the autoPET-III winning weights and
extended from two to four input channels, with the two scribble channels
zero-initialised so that the pretrained representation is preserved exactly at
initialisation. Every model is fine-tuned per fold from the corresponding
autoPET-III fold checkpoint, so that no validation case is seen during
pretraining. PET intensities are normalised against a per-scan aorta
blood-pool reference derived from a CT segmentation, which removes
tracer- and centre-specific scaling without requiring lesion labels. At
inference the five fold models are ensembled by averaging their softmax
outputs per sliding-window patch, before Gaussian-weighted stitching. On the challenge's five-fold split, with each fold
evaluated on its own validation cases, mean Dice is 0.554 and mean lesion-level
F1 is 0.528 without scribbles, rising to 0.751 and 0.733 after five correction
rounds. About 85\% of that gain follows the first scribble, and the spread between
fold models narrows five-fold over the same rounds, so interaction largely
compensates for how well or badly a given model segments unaided.

\end{abstract}
\section{Introduction}
Positron emission tomography combined with computed tomography (PET/CT) is
central to oncological staging and therapy response assessment, and
quantitative read-out of tumour burden requires lesion segmentation. Manual
delineation of whole-body studies is slow and subject to inter-reader
variation, which motivates automated methods.

The autoPET challenge series~\cite{ref_challenge}, built on a publicly released
whole-body PET/CT cohort with manual lesion annotations~\cite{ref_db_sdata,ref_db_fdg,ref_db_psma},
has driven progress on this task. Its fifth
edition (autoPET/CT V) differs from the earlier editions in two respects.
First, it is \emph{multitracer}: the cohort contains both
$^{18}$F-FDG and $^{68}$Ga/$^{18}$F-PSMA studies, whose physiological uptake
patterns differ substantially, so that a model tuned to one tracer does not
transfer to the other without care. Second, and centrally, the task is
\emph{interactive}: for each study the challenge supplies user scribbles that
mark lesion (foreground) and non-lesion (background) voxels, and the
algorithm is expected to use them. The clinically relevant question is
therefore not only how well a model segments unaided, but how reliably it
responds when a reader points at what it got wrong.

This paper describes our submitted algorithm. We treat the scribbles as two
additional input channels to a three-dimensional residual encoder U-Net,
initialise from the autoPET-III winning solution of Rokuss~et~al.~\cite{ref_autopet3},
and fine-tune on the challenge data. We report five-fold cross-validation results over successive
interactive correction rounds, which show that segmentation quality improves
consistently as scribbles accumulate, most of the gain arriving with the first
scribble.

\section{Methods}

\subsection{Data}
\subsubsection{Training data}
We use exclusively the training data provided by the challenge: 1611
whole-body PET/CT studies~\cite{ref_db_sdata,ref_db_fdg,ref_db_psma}, comprising 1014
$^{18}$F-FDG and 597 PSMA examinations; the PSMA cohort combines 369 studies
acquired with an $^{18}$F-labelled and 228 with a $^{68}$Ga-labelled PSMA
ligand; we use the v2 data release, which corrects a small number of
segmentation masks relative to the initial release. Each study is accompanied
by a manual lesion annotation. No external
or private data is used. For each study the challenge additionally provides
foreground and background scribbles; these are used as supplied and are
described in Section~\ref{sec:pre}.

\subsubsection{Validation data}
We follow the five-fold split distributed with the challenge data in
\texttt{splits\_final.json}, which divides the 1611 studies into folds of
1288/323 and 1289/322 training/validation cases. All results in
Section~\ref{sec:results} are obtained by evaluating each fold's model on that
same fold's validation split, so that every one of the 1611 studies is
predicted exactly once by a model that did not see it during training. The
same split is used by the autoPET-III reference solution of
Rokuss~et~al.~\cite{ref_autopet3}, from which our weights are initialised, and
we initialise each fold from that solution's checkpoint for the same fold;
consequently no validation case is seen during pretraining either. The
per-fold checkpoints are those published by the authors via
Zenodo~\cite{ref_autopet3_code}.

\subsection{Data pre-processing}\label{sec:pre}
All channels are resampled onto one common grid, defined from the original CT
geometry: the voxel spacing is set to $2.0364\times2.0364\times3.0$~mm and the
grid size to $\mathrm{round}(\text{size}\cdot\text{spacing}/\text{target
spacing})$ per axis, retaining the CT origin and direction. The identical
procedure is applied at training time and inside the inference container.

\subsubsection{CT.} Intensities are clipped to
$[-811.73, 1137.42]$~HU and z-scored with the dataset-wide statistics
$\mu = 124.578$, $\sigma = 280.883$, then resampled with third-order B-spline
interpolation and stored as \texttt{float16}.

\subsubsection{PET.} Standardised uptake values are divided by a per-scan
reference and passed through an inverse hyperbolic sine before a fixed
dataset-wide z-score with $\mu = 0.0904$, $\sigma = 0.2538$:
\[
\mathrm{PET}_{\mathrm{norm}}
= \bigl(\operatorname{arcsinh}(\mathrm{SUV}/(r+\varepsilon)) - \mu\bigr)/\sigma ,
\qquad \varepsilon = 10^{-8}.
\]
The reference $r$ is the robust blood-pool activity of the aorta. An aortic
mask is obtained by running TotalSegmentator~\cite{ref_totalseg} (\texttt{task="total"},
\texttt{fast=True}) on the CT, transferring the mask to the PET grid by
nearest-neighbour interpolation and eroding it by one voxel to suppress
partial-volume contamination from the vessel wall. Within the mask, focal
outliers are removed at $\mathrm{median} + 3 \cdot 1.4826 \cdot \mathrm{MAD}$
and $r$ is the mean of the remaining voxels. If the eroded mask contains fewer
than 100 voxels with positive uptake, the reference falls back to the median of
soft tissue ($-100 < \mathrm{HU} < 100$), and failing that to the median over all
positive PET voxels. This makes the normalisation label-free and
tracer-agnostic. The normalised volume is resampled with B-spline interpolation
and stored as \texttt{float16}.

\subsubsection{Scribbles.} The scribble heat maps are used exactly as provided
by the challenge and are not regenerated. In the reference implementation they
are produced by sampling up to $k=5$ connected components of the annotation
(26-connectivity). For each sampled component the single axial slice in which it
occupies the most voxels is selected, and one scribble is drawn on that slice
with the default centreline strategy, one of three strategies in the reference
simulator, whose designs follow ScribblePrompt~\cite{ref_scribbleprompt} and
nnInteractive~\cite{ref_nninteractive}: the two-dimensional slice mask is
skeletonised, the largest skeleton component is retained (8-connectivity), the
longest path between two skeleton endpoints is extracted and truncated by a
fraction of $0.1$. Background scribbles are drawn analogously inside a thin
shell obtained by twice dilating the component with a unit-radius ball and
subtracting the component itself -- a two-voxel ring hugging the lesion
boundary. At evaluation time background scribbles instead mark false-positive
regions of the prediction, so the training and test semantics of the
background channel differ; the interactive results in
Section~\ref{sec:results} measure the net effect of this mismatch. Each scribble is therefore a planar curve lying in one axial
slice, not a three-dimensional structure. The resulting
scribble voxels are written into a volume on the PET grid with value $1$ and
convolved with a Gaussian of $\sigma = 0$, which is an identity operation;
the maps are therefore \emph{binary point maps} rather than smoothed heat
maps, and are extremely sparse (of the order of $10^{-6}$ of the volume by
occupancy). Foreground and background scribbles are stored as two separate
volumes. We resample them to the target spacing with linear interpolation,
clip to $[0,1]$, and quantise to \texttt{uint8} as $\lfloor 255 x \rceil$;
they are rescaled to $[0,1]$ by division by 255 when a patch is loaded.

\subsubsection{Annotations.} Lesion masks are resampled to the target spacing
by nearest-neighbour interpolation.

\subsection{Algorithm/model}
The network is a three-dimensional residual encoder U-Net following the
nnU-Net~\cite{ref_nnunet} ResEncL preset, configured with four input channels, two output
classes and deep supervision enabled. The four channels are CT, normalised
PET, the foreground scribble map and the background scribble map, in that
order.

Weights are initialised from the autoPET-III winning solution of
Rokuss~et~al.~\cite{ref_autopet3}, which is itself pretrained on a large
multi-dataset corpus (\texttt{Dataset619\_nativemultistem}) and then fine-tuned
on the autoPET-III cohort. Two adaptations are made. First, the input stem is
extended from the two channels of the original model (CT and PET) to four; the
convolution weights of the two additional scribble channels are
\emph{zero-initialised}, so that at initialisation the extended network
computes exactly the function of the pretrained two-channel model and the
scribble pathway is learned from scratch during fine-tuning. Second, the
pretrained model carries an auxiliary organ segmentation head, which we
disable during fine-tuning: the organ information is assumed to be retained in
the encoder representation and is not supervised further. Fold $i$ of our
model is initialised from fold $i$ of the pretrained model, so the fold
structure is preserved end to end.

\subsection{Data post-processing}
Inference is performed by sliding a $192\times192\times192$ window over the
resampled volume with an overlap of $0.5$, i.e. a stride of $96$ voxels along
each axis, with the last window on each axis shifted inward so that it fits
inside the volume. Should an axis be shorter than the patch, the volume is
zero-padded at its far end and the result cropped back afterwards. The forward
passes run in mixed precision, as during training.

The per-patch class posteriors, obtained by a softmax over the network logits,
are accumulated into the output volume with a Gaussian importance weight, which down-weights the poorly contextualised patch borders
relative to the patch centre. The weight map is obtained by placing a unit
impulse at the patch centre, convolving it with a Gaussian whose standard
deviation is $\sigma_{\mathrm{scale}} = 0.125$ of the patch size along each
axis (i.e. $24$ voxels for a patch of $192$), normalising by its maximum and
clipping from below at $10^{-3}$. Two accumulators are maintained: the sum of
weighted class probabilities and the sum of the weights themselves; their
element-wise ratio yields the final probability volume, which is correct
irrespective of how many windows cover a given voxel.

The foreground channel of this probability volume is then resampled back to
the original acquisition grid by linear interpolation and clipped to $[0,1]$,
and only afterwards thresholded at $0.65$ to produce the binary lesion mask,
which is written on the reference volume's grid and affine. Applying the threshold after rather than before resampling
avoids nearest-neighbour interpolation of a binary mask, which would otherwise
discard small lesions. The threshold was selected by a sweep on fold~0's
validation split and applied unchanged to all folds and to the ensemble. No
connected-component filtering, size thresholding or other morphological
post-processing is applied.

\subsection{Training and test parameters}
Each fold is trained for 500 epochs
with stochastic gradient descent at a learning rate of $10^{-3}$, Nesterov
momentum of $0.99$, zero dampening and a weight decay of $3\times10^{-5}$. No
learning rate schedule is used. The batch size is $2$, training runs in mixed
precision, and gradients are clipped to a maximum norm of $12$.

The loss is the sum of cross-entropy and a soft Tversky term,
\[
\mathcal{L} = \mathrm{CE} + \Bigl(1 - \tfrac{1}{|C|}\textstyle\sum_{c \in C}
\tfrac{\mathrm{TP}_c + \epsilon}
{\mathrm{TP}_c + \alpha\,\mathrm{FP}_c + \beta\,\mathrm{FN}_c + \epsilon}\Bigr),
\]
with $\alpha = 0.3$, $\beta = 0.7$ and $\epsilon = 10^{-5}$. The Tversky term
is averaged over foreground classes only; including the background would
cancel the asymmetry between $\alpha$ and $\beta$, since a foreground false
negative is a background false positive. Penalising false negatives more
heavily than false positives reflects the clinical cost of missing a lesion.
The background remains supervised through the cross-entropy term. Deep
supervision is applied with the standard nnU-Net weighting
$[1, \tfrac{1}{2}, \tfrac{1}{4}, \dots]$, normalised to unit sum and with the
coarsest resolution level dropped.

Training patches are sampled stochastically rather than augmented. For a
tumour-bearing case, with probability $0.33$ a hard-negative, tumour-free
window is drawn (the first such window found within 20 attempts, otherwise the
window containing the fewest lesion voxels); otherwise a lesion is drawn
\emph{uniformly at the lesion level}, independently of its size, and a window
is placed at random such that it contains a random voxel of that lesion.
Tumour-free cases receive a uniformly random window. No intensity or geometric
augmentation of any kind is applied.

For every fold the checkpoint with the lowest validation loss over the 500
epochs is selected, using the same criterion for all five folds. Training was performed on a single NVIDIA V100
with 32~GB of memory. At test time no test-time augmentation is used. The five
fold models are ensembled by averaging their softmax outputs
\emph{per sliding-window patch}, before the patch is stitched into the output
volume, so that the Gaussian weighting and reverse resampling described above
operate on a single already-ensembled probability map.

\subsection{Github repository}
Link to Github repository: \url{https://github.com/amged-elmasry-bs/bs_autopetV_submission}

\section{Results}\label{sec:results}
We report five-fold cross-validation on the challenge training data using the
\texttt{splits\_final.json} split, evaluating each fold's model on that fold's
validation cases. All scores are produced by the challenge's official
\texttt{MetricEvaluator}~\cite{ref_autopetv_code}, using its two metrics
unmodified.

The first is the volumetric Dice between the predicted mask $P$ and the
annotation $G$,
\[
\mathrm{Dice} = \frac{2\,|P \cap G|}{|P| + |G|} .
\]
It is evaluated on the original acquisition grid. When the annotation contains no
lesion, $|G| = 0$, the quotient is undefined and the evaluator returns
\texttt{NaN}. This depends on the annotation alone: a lesion-free study yields
\texttt{NaN} whether the model correctly predicts nothing or produces a false
positive, so such cases carry no Dice at all rather than a score of zero or one.

The second is a lesion-level $F_1$. Both volumes are decomposed into connected
components at 18-connectivity, and a predicted component is counted as a true
positive when its intersection over union with an annotated lesion is at least
$0.1$; unmatched predicted components are false positives and unmatched
annotated lesions are false negatives. With those \emph{lesion counts},
\[
F_1 = \frac{2\,\mathrm{TP}}{2\,\mathrm{TP} + \mathrm{FP} + \mathrm{FN}} .
\]
Every lesion contributes equally regardless of its size, and the $0.1$ overlap
requirement means the metric asks whether a lesion was detected rather than how
precisely it was delineated. $F_1$ is likewise undefined when the annotation
contains no lesion, since $\mathrm{TP} + \mathrm{FN} = 0$.

Studies without annotated lesions are therefore excluded from both averages,
leaving 1040 of the 1611 studies scored, between 199 and 215 per fold. Two further conventions matter
for comparability. First, alongside the per-case macro average reported here,
the evaluator pools lesion counts across \emph{all} cases into an aggregated
$F_1$, in which false positives on tumour-free studies do count. Second, the
challenge ranks by the area under each metric's six-round curve, a trapezoid over the
round index in which the endpoint rounds $r_0$ and $r_5$ carry half the weight of the
interior ones. We report it normalised by the number of intervals, so that it lies on the
same $[0,1]$ scale as the metric it integrates; per-fold AUCs are given in
Tables~\ref{tab:dice} and~\ref{tab:f1}. The reported figures follow the
same convention as the official challenge leaderboard, which reports a count of
scored cases smaller than the number submitted for exactly this reason. Note that these are single-model results: each fold model is scored only on the
cases it did not train on. The five-fold ensemble that constitutes the actual
submission cannot be evaluated in this protocol, since for any training case
four of its five members have seen that case; its performance is measured on
the held-out challenge test set.

Performance is reported over successive interactive rounds. Round $r_0$ is the
unaided prediction, with both scribble channels empty. In each subsequent round
one scribble is generated from the model's current errors and appended to the
accumulated input, following the challenge's reference interactive loop: a
candidate scribble is drawn on the false-positive region ($\hat{y}=1$, $y=0$)
and another on the false-negative region ($\hat{y}=0$, $y=1$), both by the
challenge's own scribble generator. For each error region it examines every
axial slice, takes the largest two-dimensional connected component
(8-connectivity) within each, and selects the slice whose component is largest
overall; the scribble is the centreline of that component and lies entirely in
that one slice. Of the two candidates the \emph{longer} is kept -- a foreground
scribble when the false-negative scribble is at least as long as the
false-positive one, and a background scribble otherwise. A round therefore adds
a single planar curve, typically a few voxels in extent, to whichever error the
generator could trace furthest.
Scribbles are encoded exactly as at training time and as in the submitted
container: written as ones on the original grid, resampled linearly to the
target grid, clipped and quantised through \texttt{uint8}, with no additional
smoothing or renormalisation.

Each case therefore receives \textbf{six inference passes in total}: $r_0$ without any
scribble, followed by five refinement rounds $r_1 \dots r_5$, each adding one
further scribble to those already accumulated.

\begin{table}[ht]
\caption{Five-fold cross-validation, Dice over the six interactive rounds ($r_0$ without scribbles, then five refinement rounds). Each fold's model is evaluated on its own validation split. Both the per-round means and the AUC are computed over tumour-bearing cases only, since Dice is undefined on a lesion-free annotation and the evaluator returns \texttt{NaN}. The AUC is the trapezoid of the six-round curve over the round index, normalised by the number of intervals so that it shares the scale of the metric itself and its maximum is $1.0$. The Mean row averages the five fold curves.}\label{tab:dice}
\begin{tabular}{lccccccc}
\toprule
\textbf{Fold} & $r_0$ & $r_1$ & $r_2$ & $r_3$ & $r_4$ & $r_5$ & AUC \\
\midrule
0    & 0.5708 & 0.7344 & 0.7536 & 0.7591 & 0.7637 & 0.7654 & 0.7358 \\
1    & 0.4583 & 0.7130 & 0.7424 & 0.7523 & 0.7598 & 0.7638 & 0.7157 \\
2    & 0.5374 & 0.7164 & 0.7242 & 0.7281 & 0.7300 & 0.7310 & 0.7066 \\
3    & 0.6312 & 0.7291 & 0.7422 & 0.7480 & 0.7528 & 0.7540 & 0.7330 \\
4    & 0.5719 & 0.7186 & 0.7328 & 0.7390 & 0.7414 & 0.7416 & 0.7177 \\
\midrule
Mean & 0.5539 & 0.7223 & 0.7391 & 0.7453 & 0.7495 & 0.7512 & 0.7218 \\
\bottomrule
\end{tabular}
\end{table}

\begin{table}[ht]
\caption{Five-fold cross-validation, lesion-level $F_1$ over the six interactive rounds. As in Table~\ref{tab:dice}, the means and the AUC are computed over tumour-bearing cases only, $F_1$ being undefined when the annotation contains no lesion, and the AUC is the normalised trapezoid over the round index (maximum $1.0$).}\label{tab:f1}
\begin{tabular}{lccccccc}
\toprule
\textbf{Fold} & $r_0$ & $r_1$ & $r_2$ & $r_3$ & $r_4$ & $r_5$ & AUC \\
\midrule
0    & 0.5327 & 0.7160 & 0.7359 & 0.7388 & 0.7493 & 0.7506 & 0.7163 \\
1    & 0.4394 & 0.6908 & 0.7139 & 0.7232 & 0.7302 & 0.7395 & 0.6895 \\
2    & 0.5103 & 0.7010 & 0.7066 & 0.7093 & 0.7122 & 0.7135 & 0.6882 \\
3    & 0.6142 & 0.7062 & 0.7250 & 0.7272 & 0.7334 & 0.7347 & 0.7133 \\
4    & 0.5437 & 0.7039 & 0.7178 & 0.7230 & 0.7224 & 0.7246 & 0.7002 \\
\midrule
Mean & 0.5281 & 0.7036 & 0.7198 & 0.7243 & 0.7295 & 0.7326 & 0.7015 \\
\bottomrule
\end{tabular}
\end{table}

Averaged over the five folds, the unaided model reaches a Dice of $0.5539$ and a
lesion-level $F_1$ of $0.5281$. A single scribble raises these to $0.7223$ and
$0.7036$, and after five rounds they stand at $0.7512$ and $0.7326$ -- a total gain
of $+0.1973$ Dice and $+0.2045$ $F_1$. The gain is heavily front-loaded: the first
scribble alone accounts for $85\%$ of it in both metrics, with the remaining four
rounds contributing $+0.0289$ Dice and $+0.0290$ $F_1$ between them. The curves are
therefore steep at $r_1$ and close to flat by $r_3$, which is what the normalised AUC of
$0.7218$ (Dice) and $0.7015$ ($F_1$) reflects.

Dice increases at every round on all five folds. $F_1$ does so on four of them;
fold~4 dips by $0.0006$ between $r_3$ and $r_4$ before recovering, an excursion far
smaller than the between-fold spread and the only non-monotonic step in the sixty
transitions recorded. At the level of individual studies, $497$ of the $1040$ scored
cases improve in $F_1$ from $r_0$ to $r_5$, $461$ are unchanged and $82$ deteriorate.

The most pronounced effect is on the spread between folds. Unaided, the folds differ
by $0.1729$ Dice ($0.4583$ on fold~1 to $0.6312$ on fold~3); after five rounds that
spread is $0.0344$ ($0.7310$ to $0.7654$), a five-fold compression, with the
corresponding $F_1$ spread falling from $0.1748$ to $0.0371$. The ordering is not
preserved either: fold~1 is the weakest fold unaided and the second strongest at
$r_5$, whereas fold~2 -- whose selected checkpoint also has the highest validation
loss of the five -- gains least after the first scribble and ends last. Interaction
thus compensates for much of the variation in unaided quality between fold models,
which is the behaviour one wants from a corrective interface.

\section{Discussion}
The cross-validation results in Tables~\ref{tab:dice} and~\ref{tab:f1} show
that segmentation quality increases with each interactive round: the model
follows the supplied scribbles and progressively corrects its own errors
rather than ignoring the additional channels or degrading as more input is
accumulated.

Averaged over folds the loop is worth $+0.1973$ Dice and $+0.2045$ $F_1$, and the
first scribble supplies roughly $85\%$ of that; the four subsequent rounds refine
rather than transform the segmentation.

Where the gain comes from is worth separating out. Pooled over all folds and cases,
five rounds convert $829$ false negatives into true positives, while the false
positive count rises slightly, from $6869$ to $6998$. The improvement is thus almost
entirely a matter of recovering lesions the model had missed, not of suppressing
spurious ones. This is despite the loop spending most of its budget on the latter:
under the challenge's rule the longer of the two candidate scribbles is kept, and
because false-positive regions are typically larger and more elongated than the
residual false-negative ones, $69\%$ of the $5426$ scribbles drawn are background
scribbles. Those background scribbles are, on this evidence, close to inert. The
likely reason is the semantic mismatch noted in Section~\ref{sec:pre}: during
training the background channel carries a thin shell hugging a true lesion, marking
its boundary, whereas at evaluation it marks a whole false-positive region, and the
network has never been trained to read it that way. Training with evaluation-style
background scribbles -- drawn on the model's own false positives rather than around
annotated lesions -- is the most promising single change suggested by these results.

Two design choices are worth noting. Normalising PET against a per-scan aorta
blood-pool reference is label-free and does not depend on the tracer, which
matters for a cohort combining FDG and PSMA studies whose absolute uptake
scales differ. Zero-initialising the two scribble channels means the extended
four-channel network reproduces the pretrained two-channel model exactly
before any fine-tuning, so the interactive capability is added without
disturbing the segmentation behaviour inherited from the autoPET-III solution.

Several limitations should be stated. The cross-validation figures are
single-model results; the submitted five-fold ensemble cannot be scored under this
protocol, so its benefit is assumed from the usual behaviour of nnU-Net ensembles
rather than measured here. The threshold of $0.65$ was chosen on fold~0 and reused
unchanged everywhere, including for the ensemble, whose probability distribution is
not identical to a single model's. Scores are reported on tumour-bearing studies
only, because both metrics are undefined without an annotated lesion; this excludes
$571$ of the $1611$ studies and means that false positives on lesion-free scans,
clinically the more costly error, do not enter the per-case averages at all. The
simulated reader is also an optimistic one: scribbles are generated automatically
from the ground truth by the challenge's own simulator, always mark a genuine error,
and are drawn as clean centrelines, so the results bound what a perfectly informed
annotator would obtain rather than what a human would. Finally, the background
channel behaves as described above, which caps what further rounds can achieve.

\section{Conclusion}
We described our submission to the autoPET/CT V challenge: a
scribble-conditioned three-dimensional residual encoder U-Net, initialised
per-fold from the autoPET-III winning weights, extended from two to four input
channels with zero-initialised scribble pathways, trained with a
false-negative-weighted Tversky and cross-entropy objective, and deployed as a
five-fold ensemble whose fold predictions are averaged per sliding-window
patch. Five-fold cross-validation shows that performance improves
consistently across successive interactive rounds, confirming that the model
uses the supplied scribbles to correct its own errors. Averaged over the five folds, Dice rises from $0.5539$ unaided to $0.7512$ after
five scribbles and lesion-level $F_1$ from $0.5281$ to $0.7326$, with about $85\%$ of
the gain delivered by the first scribble, and the spread between fold models narrows
five-fold over the same rounds.

\begin{table}[ht]
\caption{Algorithm details}\label{tab1}

\begin{tabular}{P{0.17\textwidth}P{0.15\textwidth}P{0.24\textwidth}P{0.21\textwidth}P{0.15\textwidth}}
\toprule
\textbf{Team name} & \textbf{algorithm name} & \textbf{data pre-processing} & \textbf{data post-processing} & \textbf{training data augmentation} \\
\midrule
BS &  BS baseline &
Resampled to a common grid derived from the CT geometry at
2.0364 x 2.0364 x 3.0 mm. CT: clip to $[-811.7, 1137.4]$ HU, dataset z-score,
B-spline. PET: $\operatorname{arcsinh}$(SUV / per-scan aorta blood-pool
reference), dataset z-score, B-spline; reference from a TotalSegmentator aorta
mask with MAD outlier rejection. Scribbles: linear resample, clip $[0,1]$,
\texttt{uint8}. &
Sliding window $192^3$ at $0.5$ overlap; per-patch softmax stitched with
Gaussian importance weights ($\sigma = 0.125 \times$ patch); reverse linear
resample to the original grid; threshold $0.65$. No connected-component or
size filtering. &
None. Stochastic patch sampling only ($p_{\mathrm{bg}}=0.33$ hard negatives,
otherwise per-lesion uniform) \\
\bottomrule
\end{tabular}

\vspace{2em}

\begin{tabular}{P{0.12\textwidth}P{0.20\textwidth}P{0.16\textwidth}P{0.22\textwidth}P{0.20\textwidth}}
\toprule
\textbf{test time augmentation} & \textbf{ensembling} & \textbf{standardized framework} & \textbf{network architecture} & \textbf{loss} \\
\midrule
None &
5-fold cross-validation ensemble; softmax averaged per sliding-window patch,
before Gaussian stitching &
nnU-Net, ResEncL preset (3D) &
Residual Encoder U-Net (3D); 4 input channels (CT, PET, foreground and
background scribbles), 2 output classes, deep supervision &
Cross-entropy $+$ Tversky ($\alpha=0.3$, $\beta=0.7$, foreground classes only),
deep-supervision weighted \\
\bottomrule
\end{tabular}

\vspace{2em}

\begin{tabular}{P{0.22\textwidth}P{0.20\textwidth}P{0.32\textwidth}P{0.18\textwidth}}
\toprule
\textbf{training data} & \textbf{data/model dimensionality and size} & \textbf{use of pre-trained models} & \textbf{GPU hardware for training} \\
\midrule
1014 FDG $+$ 597 PSMA PET/CT of autoPET (369 $^{18}$F- and 228 $^{68}$Ga-labelled
PSMA); 5-fold split of \texttt{splits\_final.json} &
3D; patch $192\times192\times192$ at 2.0364 x 2.0364 x 3.0 mm &
Public: autoPET-III winning model of Rokuss et al.~\cite{ref_autopet3,ref_autopet3_code},
itself pretrained on \texttt{Dataset619\_nativemultistem}. Fold $i$ initialised
from their fold $i$; input stem extended 2 $\rightarrow$ 4 channels with the two
scribble channels zero-initialised; auxiliary organ head disabled &
1x NVIDIA V100 (32 GB) \\
\bottomrule
\end{tabular}
\end{table}


\begin{credits}

\subsubsection{\discintname}
The authors have no competing interests to declare that are relevant to the
content of this article. All authors are employees of Brightskies. No author
is affiliated with any of the institutes of the challenge organizers.
\end{credits}
%
%
%

\begin{thebibliography}{8}
\bibitem{ref_autopet3}
Rokuss, M., Kovacs, B., Kirchhoff, Y., Xiao, S., Ulrich, C., Maier-Hein,
K.H., Isensee, F.: From FDG to PSMA: A Hitchhiker's Guide to Multitracer,
Multicenter Lesion Segmentation in PET/CT Imaging. arXiv:2409.09478 (2024)

\bibitem{ref_autopet3_code}
Rokuss, M.: Model checkpoint of the autoPET III LesionTracer solution.
Zenodo (2024). \doi{10.5281/zenodo.14007247}. Code:
\url{https://github.com/MIC-DKFZ/autopet-3-submission}. Last accessed 26 Aug 2026

\bibitem{ref_autopetv_code}
autoPET/CT V challenge: official evaluation code and interactive baseline.
\url{https://github.com/lab-midas/autoPETV}. Last accessed 26 Aug 2026

\bibitem{ref_nnunet}
Isensee, F., Jaeger, P.F., Kohl, S.A.A., Petersen, J., Maier-Hein, K.H.:
nnU-Net: a self-configuring method for deep learning-based biomedical image
segmentation. Nature Methods \textbf{18}(2), 203--211 (2021)

\bibitem{ref_totalseg}
Wasserthal, J., Breit, H.-C., Meyer, M.T., Pradella, M., Hinck, D., Sauter,
A.W., Heye, T., Boll, D.T., Cyriac, J., Yang, S., Bach, M., Segeroth, M.:
TotalSegmentator: Robust Segmentation of 104 Anatomic Structures in CT Images.
Radiology: Artificial Intelligence \textbf{5}(5), e230024 (2023)

\bibitem{ref_db_sdata}
Gatidis, S., Hepp, T., Fr\"uh, M., La Foug\`ere, C., Nikolaou, K., Pfannenberg,
C., Sch\"olkopf, B., K\"ustner, T., Cyran, C., Rubin, D.: A whole-body
FDG-PET/CT dataset with manually annotated tumor lesions. Scientific Data
\textbf{9}, 601 (2022). \doi{10.1038/s41597-022-01718-3}

\bibitem{ref_db_fdg}
Gatidis, S., Kuestner, T.: A whole-body FDG-PET/CT dataset with manually
annotated tumor lesions (FDG-PET-CT-Lesions) [dataset]. The Cancer
Imaging Archive (2022). \doi{10.7937/gkr0-xv29}

\bibitem{ref_db_psma}
Jeblick, K., Schachtner, B., Mittermeier, A., Dexl, J., Wesp, P., K\"ustner, T.,
Gatidis, S., Fr\"uh, M., Fabritius, M., Unterrainer, L., Sheikh, G., Delker, A.,
B\"oning, G., Brendel, M., Ricke, J., Werner, R., Gu, S., Ingrisch, M., Geyer,
T., Cyran, C.: A whole-body PSMA-PET/CT dataset with manually annotated tumor
lesions (PSMA-PET-CT-Lesions) (Version 1) [dataset]. The Cancer Imaging
Archive (2024). \doi{10.7937/r7ep-3x37}

\bibitem{ref_scribbleprompt}
Wong, H.E., Rakic, M., Guttag, J., Dalca, A.V.: ScribblePrompt: Fast and
Flexible Interactive Segmentation for Any Biomedical Image. In: ECCV 2024.
arXiv:2312.07381

\bibitem{ref_nninteractive}
Isensee, F., Rokuss, M., Kr\"amer, L., et al.: nnInteractive: Redefining 3D
Promptable Segmentation. arXiv:2503.08373 (2025)

\bibitem{ref_challenge}
Kuestner, T., Gatidis, S., Megne, O., Peisen, F., Ingrisch, M., Fabritius, M.,
Dexl, J., Jeblick, K., Cyran, C., Shiyam Sundar, L.K., Marinov, Z., Jackson, P.,
Hofman, M., Kleesiek, J., Kim, M., Herrmann, K.: AutoPET: Automated Lesion
Segmentation in Whole-Body PET/CT (v1). Zenodo (2026).
\doi{10.5281/zenodo.19714420}

\end{thebibliography}
%

\end{document}